\documentclass[10pt,journal,final,finalsubmission,twocolumn]{article}
\catcode`\@=11

\usepackage{amsmath,amssymb} 
\usepackage{color}
\usepackage[]{graphicx}

\newtheorem{definition}{Definition}

\newtheorem{theorem}{\textbf{Theorem}}

\newcommand{\tdot}[1]{\smash{\mathop{#1}\limits^{...}}}

\newcommand{\evat}[1]{\smash{\mathop{ }_{#1}}}

\newcommand{\bmath}[1]{\mbox{ \boldmath $\!#1\!$ \unboldmath}}

\usepackage{amsfonts}
\usepackage{bm}
\usepackage[small,bf]{caption}
\usepackage{multirow} 
\usepackage{rotating}
\usepackage{lineno}
\usepackage{float}
\usepackage{enumerate}

\newcommand{\eqalign}[1]{\null\,\vcenter{\openup\jot \m@th
\ialign{\strut\hfil$\displaystyle{##}$&$
\displaystyle{{}##}$\hfil \crcr#1\crcr}}\,}

\newcommand{\eqalignno}[1]{\displ@y \tabskip=\centering
\halign to\displaywidth{\hfil$\@lign\displaystyle{##}$
\tabskip=0pt &$\@lign\displaystyle{{}##}$
\hfil\tabskip=\centering
&\llap{$\@lign##$}\tabskip=0pt\crcr #1\crcr}}

\newcommand{\leqalignno}[1]{\displ@y \tabskip=\centering
\halign to\displaywidth{\hfil$\@lign\displaystyle{##}$
\tabskip=0pt &$\@lign\displaystyle{{}##}$
\hfil\tabskip=\centering &\kern-\displaywidth\rlap{$\@lign##$}
\tabskip=\displaywidth\crcr #1\crcr}}

\begin{document}
%
\title{Incremental Noising and its Fractal Behavior} 

\author{Konstantinos~A.~Raftopoulos\thanks{K.Raftopoulos, raftop@image.ntua.gr,  Computer Science Division, National Technical University of Athens, Iroon Polytechneiou 9, 15780.},~Marin~Ferecatu\thanks{M.Ferecatu, marin.ferecatu@cnam.fr, Conservatoire National des Arts et Métiers, Laboratoire CEDRIC, Equipe Vertigo, 292 rue Saint-Martin 75003 Paris~-~France.},~Dionyssios~D.~Sourlas\thanks{D.Sourlas, dionyssis.sourlas@unilever.com, Unilever Hellas SA, Seneka 10, Kifisia 145 64.},~Stefanos~D.~Kollias\thanks{S.Kollias, stefanos@cs.ntua.gr, Computer Science Division, National Technical University of Athens, Iroon Polytechneiou 9, 15780.}
}
\date{}
\maketitle

\begin{abstract}
This manuscript is about further elucidating the concept of \textit{noising}. The concept of \textit{noising} first appeared in \cite{CVPR14}, in the context of curvature estimation and vertex localization on planar shapes. There are indications that \textit{noising} can play for global methods the role smoothing plays for local methods in this task. This manuscript is about investigating this claim by extending \textit{noising} to \textit{incremental noising}, in a recursive deterministic manner, analogous to how smoothing is extended to progressive smoothing in similar tasks. As investigating the properties and behavior of \textit{incremental noising} is the purpose of this manuscript, a surprising connection between \textit{incremental noising} and \textit{progressive smoothing} is revealed by the experiments. To explain this phenomenon, the \textit{fractal} and the \textit{space filling} properties of the two methods respectively, are considered in a unifying context. 
\end{abstract}

%

\section{Introduction}

For every point on a curve one may consider the sum of its distances to all the other points. This \textit{total distance} of a point to all the other points (or to the rest of the curve as one may choose to see it) captures a relationship between \textit{location} and curvature. In \cite{CVIU} it is called the \textit{View Area Representation} (VAR) descriptor. In \cite{CVPR14} the \textit{Global Local (GL) Equations} are based on VAR to describe relations between location and curvature. 
VAR's resistance to noise has been investigated in \cite{CVIU} with various experiments. In \cite{CVPR14} it was shown that \textit{noising}, inducing, that is, additional random noise on the boundary actually helps a VAR-based representation of vertices. This concept will be extended and further investigated in this manuscript with the introduction of \textit{incremental noising}. Experimental results will guide us to surprising conclusions regarding a hidden connection between incremental noising and progressive (multi scale) smoothing.  

\section{Related Work and Contribution}
This manuscript is about further elucidating the general concept of \textit{noising} and its incremental behavior in particular. \textit{Noising} is an interesting and intriguing idea. Its various properties however, e.g. its incremental behavior, are not quite understood yet. A full potential for applications will be envisioned if we manage to understand the basic properties of noising and in this understanding we contribute with this manuscript.

In \cite{CVIU}, the \textit{VAR descriptor} is defined, and used to define curvature in a global sense whereas in \cite{CVPR14} the concept of \textit{noising} is introduced. In this paper we continue along line of research with two main contributions:

(a) The concept of noising is evolved to that of \textit{incremental noising} and its behavior is investigated in relation to vertex localization against various baseline methods at different degrees of locality and smoothing characteristics. The extension from noising to incremental noising introduces novel concepts that are related to the \textit{space filling} properties of a curve, enabling further insight into the general concept of noising.     

(b) A hidden link is discovered between the new concept of incremental noising and the one of progressive smoothing. This link is significant because two conceptually orthogonal methods are now for the first time connected, elucidating this way a unification playground by means of the space filling properties of curves. 

Related work on treating shapes as real functions and estimating curvature can be found in \cite{ssoatto,rel1,rel2,rel3,rel4,rel5,rel6,rel7,rel8,rel9,rel10,rel11,rel12,rel13,rel15}.

The rest of the paper is as follows: In the next section a connection to previous material on the VAR descriptor is provided together with the concept of nosing and its extension to incremental noising. The Experimental section follows. A discussion closes the paper.

\section{Connection to Previous Material}

\begin{definition}
 Let $\lbrack0,\lambda\rbrack\subset\mathbb{R}$ and $\bmath{\alpha} : \lbrack 0,\lambda\rbrack\rightarrow \mathbb{R} ^2$ a continuous at least $C^3$, closed {\it planar curve of length $\lambda$} in $\mathbb{R} ^2$, parametrized with respect to the arc length $s$ and $\phi_\alpha$ be a distance function defined on $\lbrack0,\lambda\rbrack$ and taking values in $\mathbb{R}$ as follows:
\begin{equation}
\eqalign{
\phi_\alpha:\lbrack0,\lambda\rbrack\rightarrow\mathbb{R}\colon s\mapsto\phi_\alpha(s):=\left.\int_0^\lambda\Vert\bmath{\alpha}(s)-\bmath{\alpha}(\xi)\Vert\right.d\xi
}
\end{equation}

\end{definition}
$\phi_\alpha (s)$ is called the \textit{VAR descriptor} and can be interpreted as modeling a notion of \textit{total distance} between the curve point $\bmath{\alpha}(s)$ and the rest of the curve. 

Now let $s_*,\in (0,\lambda ]$ such that the normal to the curve at $\bmath{\alpha }(s_*)$ is considered explicitly and $\xi\in (0,\lambda ]$ with $\xi\ne s_*$ signifying a random point $\bmath{\alpha}(\xi)$ on the curve. We denote with $\bmath{r}(s_*,\xi)\equiv\bmath{r}$ the vector $\bmath{\alpha}(s_*)-\bmath{\alpha}(\xi)$ and $\omega(s_*,\xi)\equiv\omega$ the angle from the \textit{normal} to the curve at $s_*$ to $-\bmath{r}(s_*,\xi)$ measured counter-clockwise. In the form of a Theorem, we gather results from \cite{CVIU}. Dots represent derivatives always with respect to $s$. 
 \begin{theorem}\label{propcurv}
Let $\bmath{\alpha}\in C^3( (0, \lambda\rbrack, \mathbb{R}^2)$ a closed planar curve of nonzero length $\lambda$, as above. If  $\varphi_\alpha (s)$ the total distance function (VAR descriptor), $\kappa (s)$ the curvature function and $s_*,\xi,\bmath{r}$ and $\omega$ as above, then:
\begin{enumerate}[(a)]
\item
\begin{equation}
\label{eq:L156}
\eqalign{
 \dot\varphi_\alpha (s_*)=-\left.\int_0^\lambda sin(\omega)d\xi\middle|\evat{s=s_*}\right.
}
\end{equation}
\item 
\begin{equation}
\label{eq:L15aa}
\eqalign{
\ddot\varphi_\alpha(s_*)=\kappa(s_*)A(s_*)+B(s_*)
}
\end{equation}

where  $A(s_*)=\left.\int_0^\lambda cos(\omega)d\xi\middle|\evat{s=s_*}\right.$ and $B(s_*)=\left.\int_0^\lambda\frac{cos^2(\omega)}{\Vert \bmath{r}\Vert}d\xi\middle|\evat{s=s_*}\right.$ global shape descriptors measured at $\bmath\alpha (s_*)$. 
\item 
 If in addition, $\varphi_\alpha(s_*)$ a local extremum of $\varphi_\alpha(s)$. Then $\kappa(s_*)\ne 0$ and $A(s_*)\ne 0$ and
\begin{equation}
\label{eq:L15aaa}
\eqalign{
\kappa(s_*)=\frac{\ddot\varphi_\alpha(s_*)-B(s_*)}{A(s_*)}
}
\end{equation}
\end{enumerate}
\end{theorem}
\noindent
~\\[1pt]
\noindent

\subsection{Noising with Gaussian Perturbations}\label{sec:noising}
In \cite{CVPR14} a global \textit{noising} algorithm is designed, consisting of {random high frequency perturbations} on the boundary of noisy or smooth shapes and shown to have advantages in vertex identification over state of the art local methods that don't alter the shape by smoothing.  
The method is based on equation(\ref{eq:L15aaa}) which defines curvature through global descriptors at the local extrema of $\phi$. Since all the quantities on the right hand side of equation (\ref{eq:L15aaa}) are integrals defined on the whole of the shape, they don't change significantly with noise, therefore this definition of curvature seems \textit{stronger than the traditional one}. In fact, in \cite{CVPR14}, it is demonstrated that noise not only is not affecting significantly this definition of curvature but it also improves the identification of vertices, giving rise to the concept of \textit{noising} as opposed to \textit{smoothing}. This result is counterintuitive since vertices are third order differentials, thus even more sensitive to noise than curvature is with traditional methods.
According to \cite{CVPR14} equation (\ref{eq:L15aaa}) suggests a method of identifying points of extreme location and curvature in the \textit{collocation} of $\dot\phi$ and $\tdot\phi$ zero crossings. Under this method noise would have no effect since it doesn't affect location significantly. In fact, the experiments in \cite{CVPR14} indicate that at points where the curve is not equally displaced around the normal, the induction of further Gaussian noise around these points has the effect of \textit{correcting} the curve's displacement, the total distance function acquiring this way better visibility in identifying maximum curvature locations through $\ddot\phi$. 

\subsection{Incremental Noising with Deterministic Recursion}\label{sec:NoiseDet}
Here we extend the method of noising to that of \textit{incremental noising} by inducing deterministic perturbations on the boundary in a recursive fashion and we will examine how well VAR-based identification of vertex points performs in this case, in relation to various local methods at different degrees of locality. This new \textit{noising} process can be performed in an additive manner to the existing boundary, also not affecting the initial boundary points. In the discrete case of a digital curve, for each pair of consecutive points on the initial boundary, a \textit{new point} is added at the intersection of the circles centered at the original points and having equal radii of a certain length, greater than half the distance between the two original points. The difference with the \textit{noising} defined in \cite{CVPR14} is that there the radii of the circles were drawn from a Gaussian distribution, whereas here the radii are constant as a parameter of the method. Furthermore here, \textit{noising} is applied recursively, doubling the curve's points at each successive step. This \textit{incremental noising}, proposed in this paper, is further investigated here for the first time in its combination with VAR for vertex identification. Different implementations involving the locations of $\dot\phi$ and $\tdot\phi$ zero-crossings are compared against local, localized and smoothing methods that employ various degrees of locality.
Incremental noising works in this case because the use of VAR-based global representation of vertices turns the \textit{negative local} effects of noise into \textit{positive global} effects. For the proposed method in particular, noising is an enabler. Vertices are detected \textit{directly} without the need for curvature calculations. Recursively applying noising in the proposed manner forms neighborhoods of increasing differential order \textit{around} the initial curve points, resulting in a concept that is analogous to that of progressive smoothing. Thus \textit{incremental noising} used with global methods, can be viewed as a conceptual duality to what \textit{progressive smoothing} is for local methods. In the next sections we further investigate these concepts.   

\section{Experimental Investigation}\label{sec:exp}

In the experimental section we seek to verify that while \textit{incremental noising} is an enabler for global methods it is an inhibitor for local methods. In other words we seek to differentiate between global and local methods based on the effect \textit{incremental noising} has on them. We show there is at least one important application, namely \textit{vertex localization} where this indeed happens. 

The baseline methods themselves are derived from the different ways one can define discrete curvature on a digital curve according to the fundamentals in the literature. The implementation is our own because the methods should be as clear and as simple as possible, so the reader can focus on the scope of locality employed by each method, and how \textit{incremental noising} is affecting them. The experimental results on the effects of \textit{incremental noising} on vertex localization are better understood this way. The baselines are meant to serve different definitions (implementations) of curvature so as to see how \textit{incremental noising} is affecting the task of vertex localization under these different implementations. 

A surprising result is also revealed in the experiments. The proposed VAR-based global method with incremental noising, identifies on distorted shapes, the same points that progressive smoothing identifies on the corresponding undistorted ones. Deterministic incremental noising in other words, seems to have the effect of gradually removing Gaussian distortions on the boundary, in terms of identifying vertices on the corresponding smooth shapes. This result is counterintuitive and was reached by means of precision vs recall (PR) comparisons in identifying perceptually interesting points (vertices) on the boundary of the KIMIA\cite{kimia} benchmark dataset of silhouettes for various local, localized and smoothing methods that assume different degrees of locality in their treatment of noise. We offer a unifying explanation of this connection between incremental noising and progressive smoothing.

\subsection{Experimental Design}\label{sec:ED}

The KIMIA dataset consists of 9 classes of 11 shapes each and has been used expensively in the literature to benchmark the performance of various classification algorithms. Here we use shapes from all classes, to produce representative \textit{noisy} versions of shapes from each class and test various methods on their ability to identify Interesting Points (IPs) on their noisy boundaries. Term \textit{noisy} is used to describe the resulting contours after Gaussian noise has been induced on the original contours. The resulting noisy contours are further processed with \textit{noising}. The term \textit{noising} is used to describe  the process of making a contour of $P$ points $P>100$ from a contour of $P/2$ points after adding new points to it in a principled manner as was explained in section \ref{sec:NoiseDet}. From each \textit{noisy} contour of $n$ points a \textit{noising} step produces another \textit{noisy} contour of $2n$ points. Similarly in a recursive manner contours of points $2^kn$ are produced for each KIMIA silhouette, $k$ being the $k^{th}$ noising step for $k=1,\dots,4$. 
Precision vs recall (PR) measurements, of the proposed method against local and localized variants of methods for identifying interesting points on the boundary of the above noisy versions of shapes are appropriately designed and performed in the following sections.
 
\newcommand*{\w}{4}
\newcommand*{\h}{6.6}
\newcommand*{\m}{4}
\begin{figure*}[]
\makebox[\textwidth][c]{
\begin{minipage}[b]{\m cm}
\includegraphics[width=\w cm, height=\h cm]{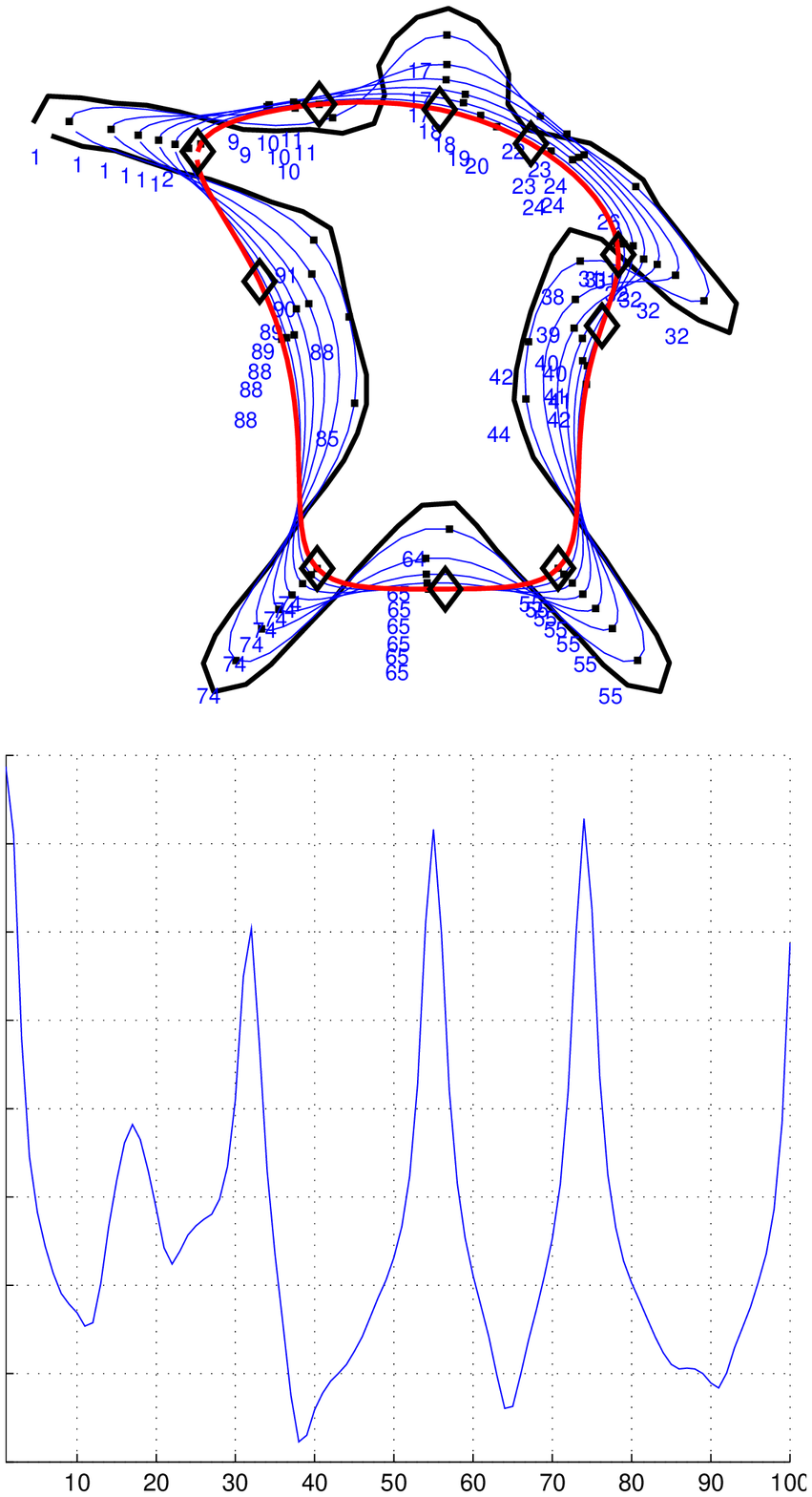}
\end{minipage}
\begin{minipage}[b]{\m cm}
\includegraphics[width=\w cm, height=\h cm]{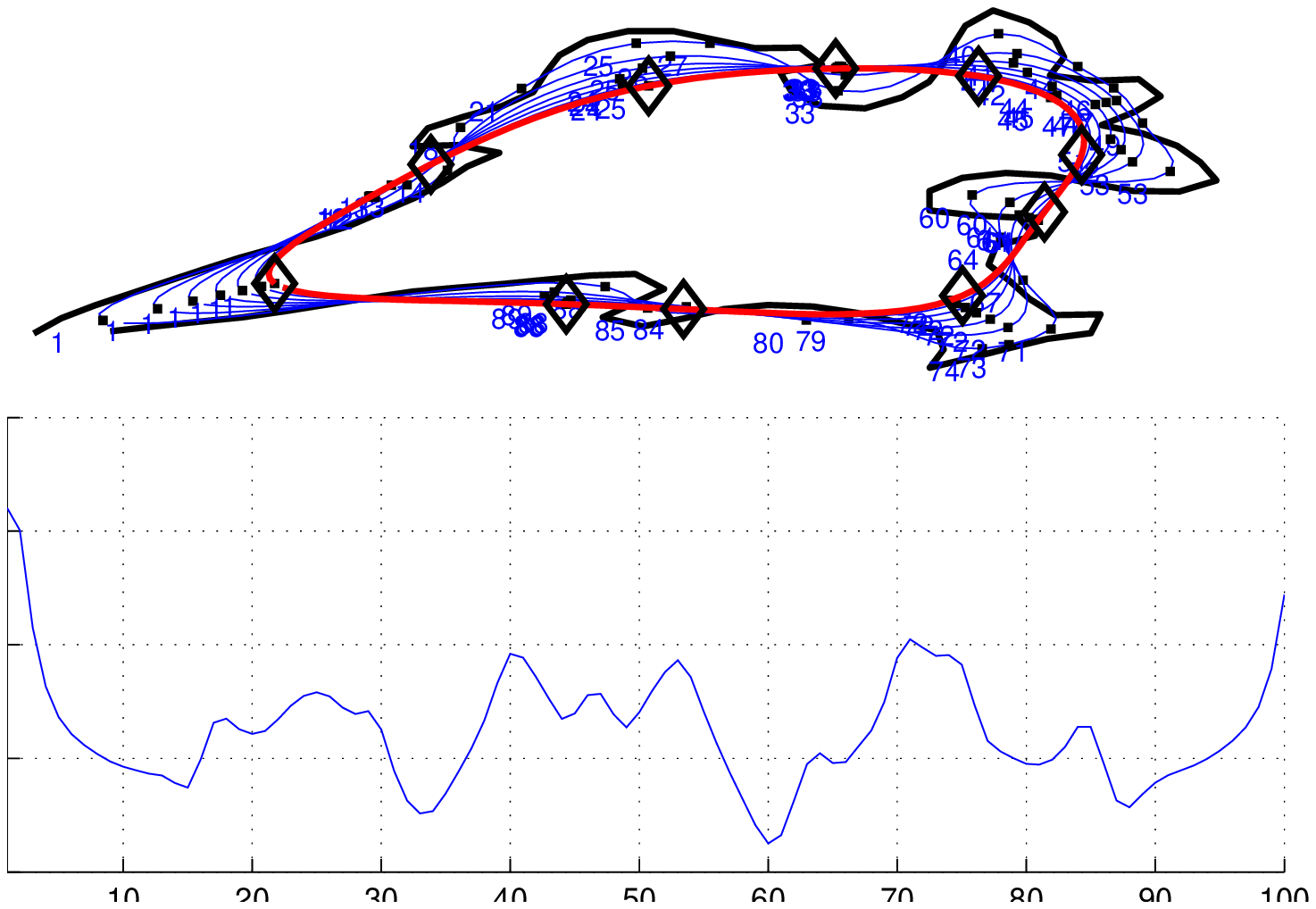}
\end{minipage}
\begin{minipage}[b]{\m cm}
\includegraphics[width=\w cm, height=\h cm]{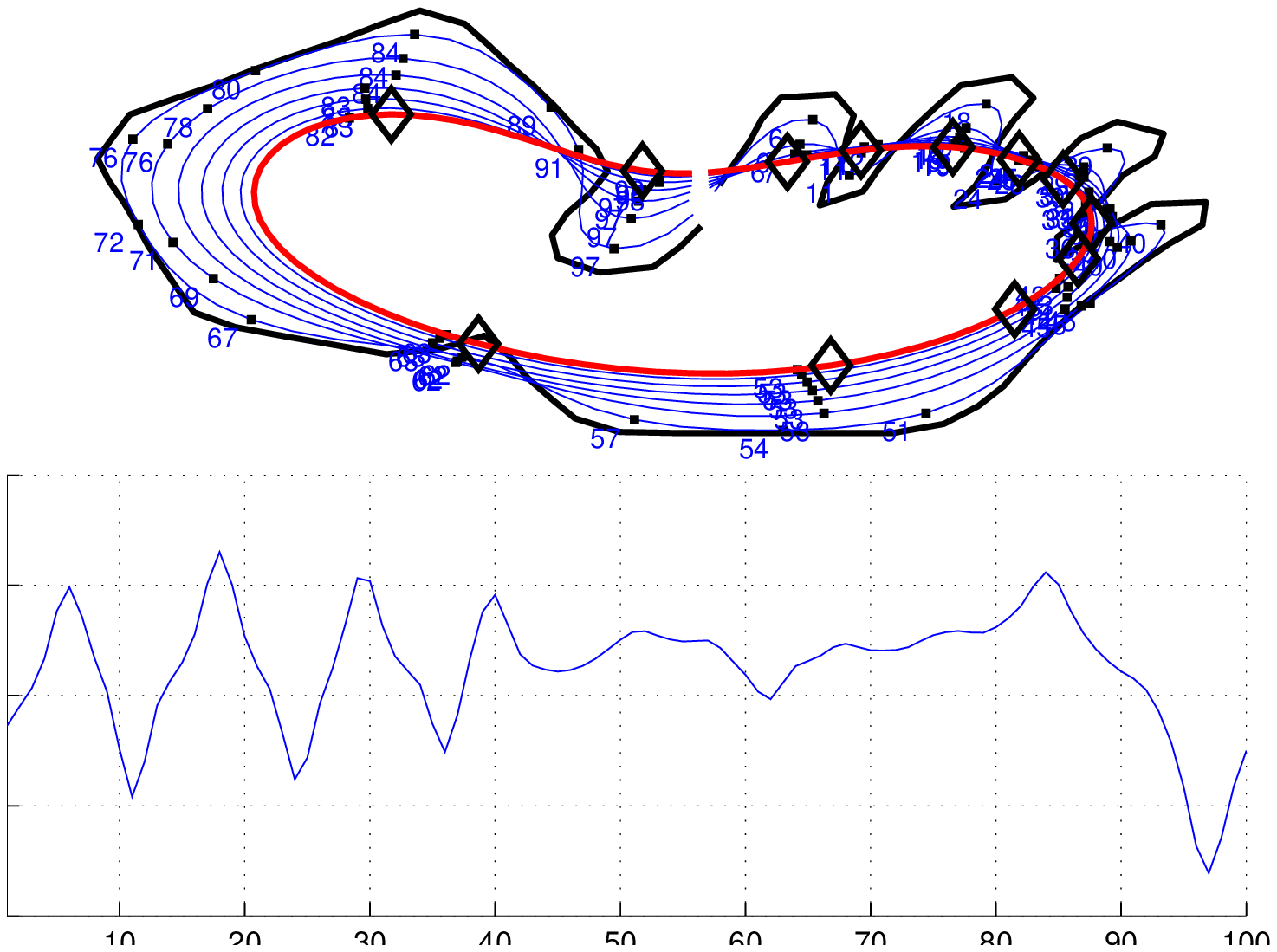}
\end{minipage}
}
\makebox[\textwidth][c]{
\begin{minipage}[b]{\m cm}
\includegraphics[width=\w cm, height=\h cm]{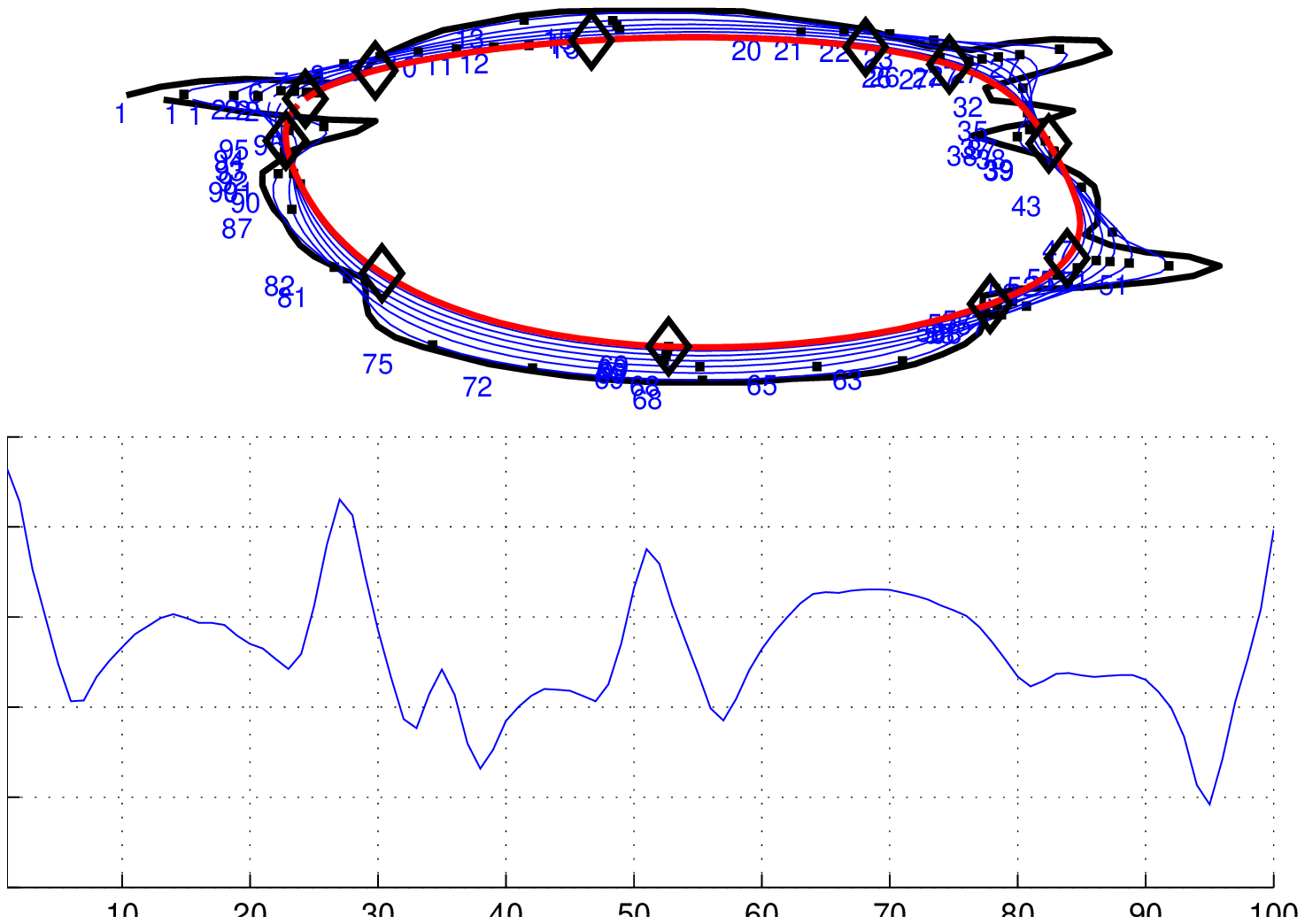}
\end{minipage}
\begin{minipage}[b]{\m cm}
\includegraphics[width=\w cm, height=\h cm]{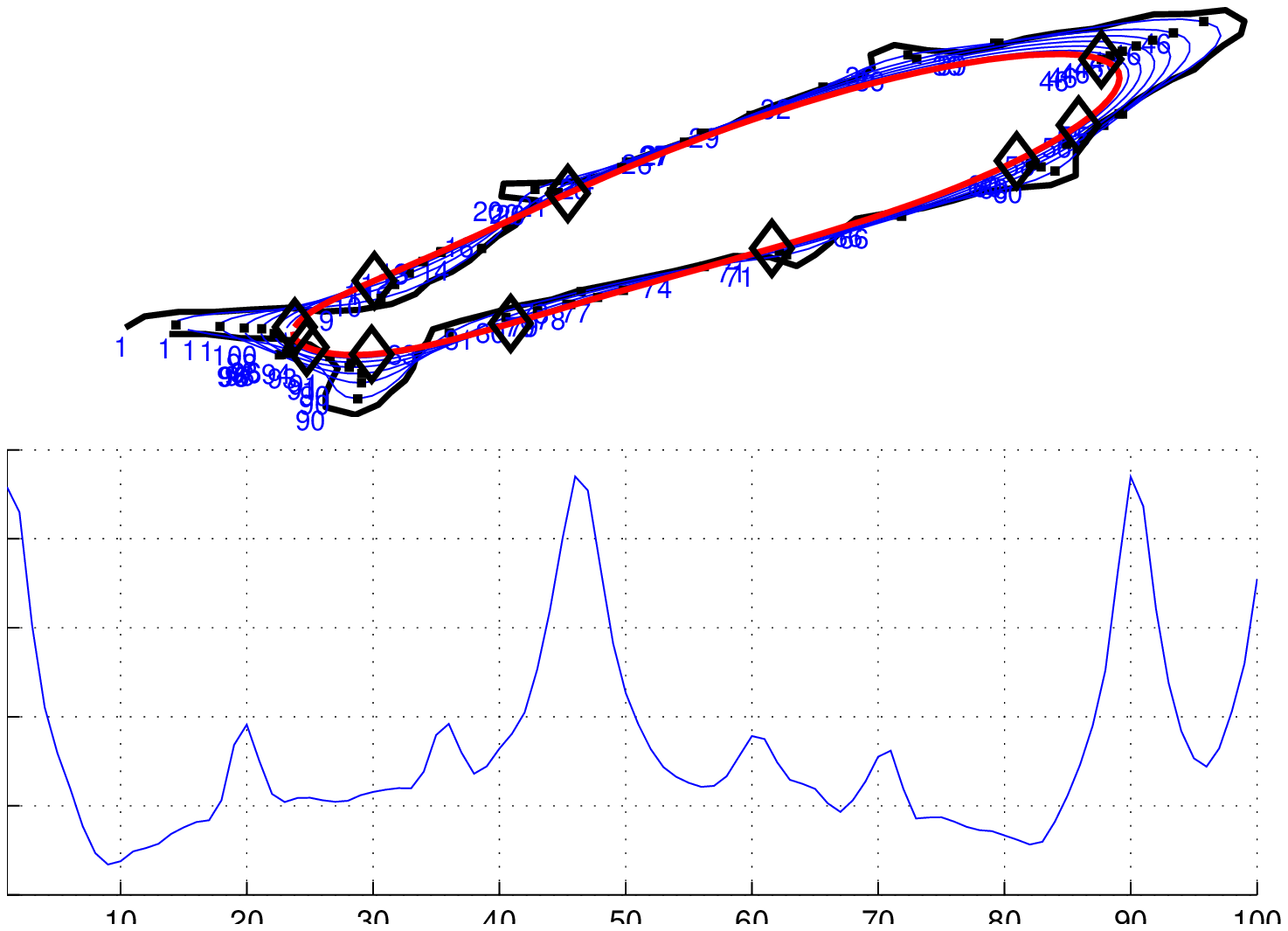}
\end{minipage}
\begin{minipage}[b]{\m cm}
\includegraphics[width=\w cm, height=\h cm]{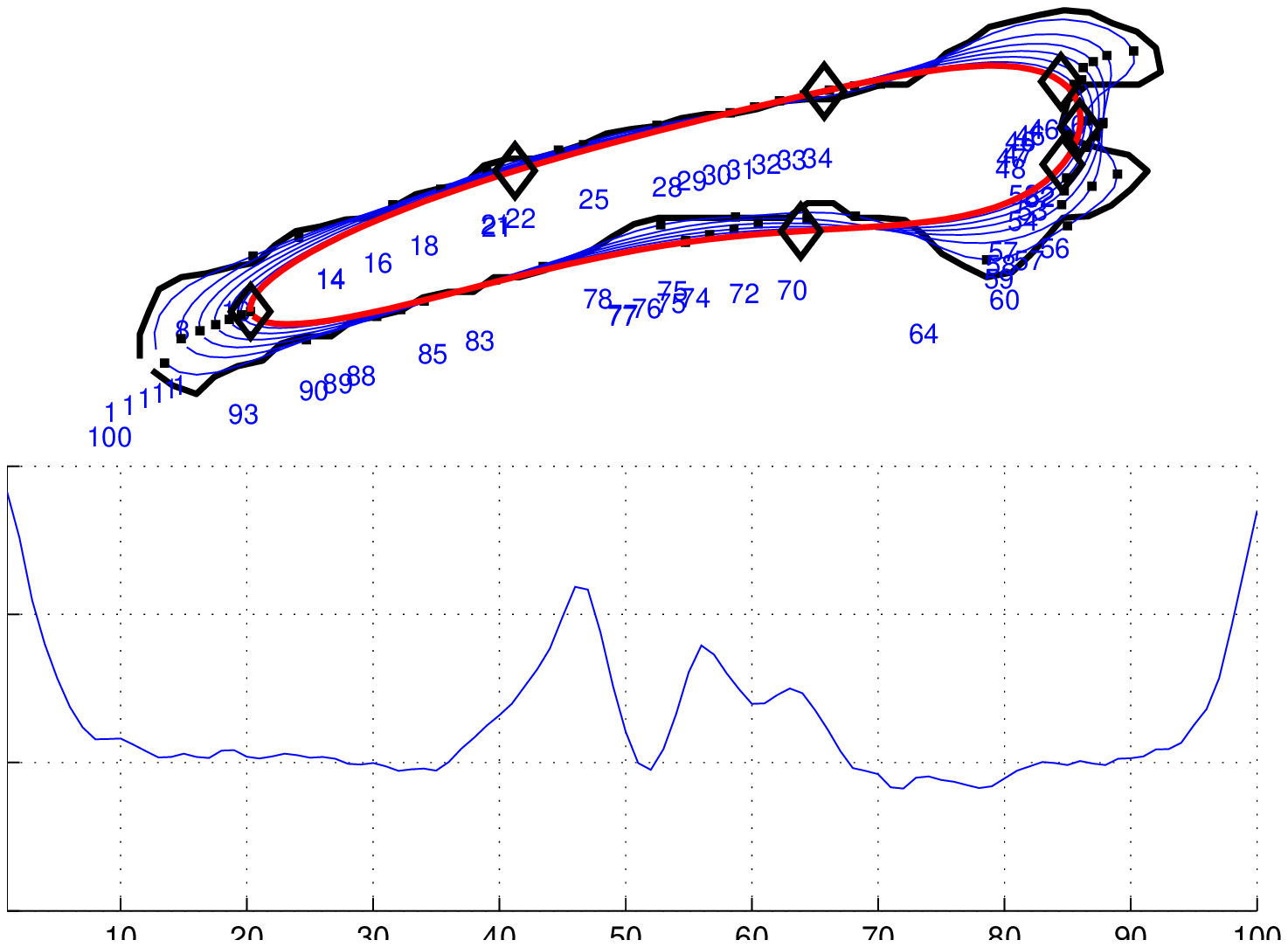}
\end{minipage}
}
\makebox[\textwidth][c]{
\begin{minipage}[b]{\m cm}
\includegraphics[width=\w cm, height=\h cm]{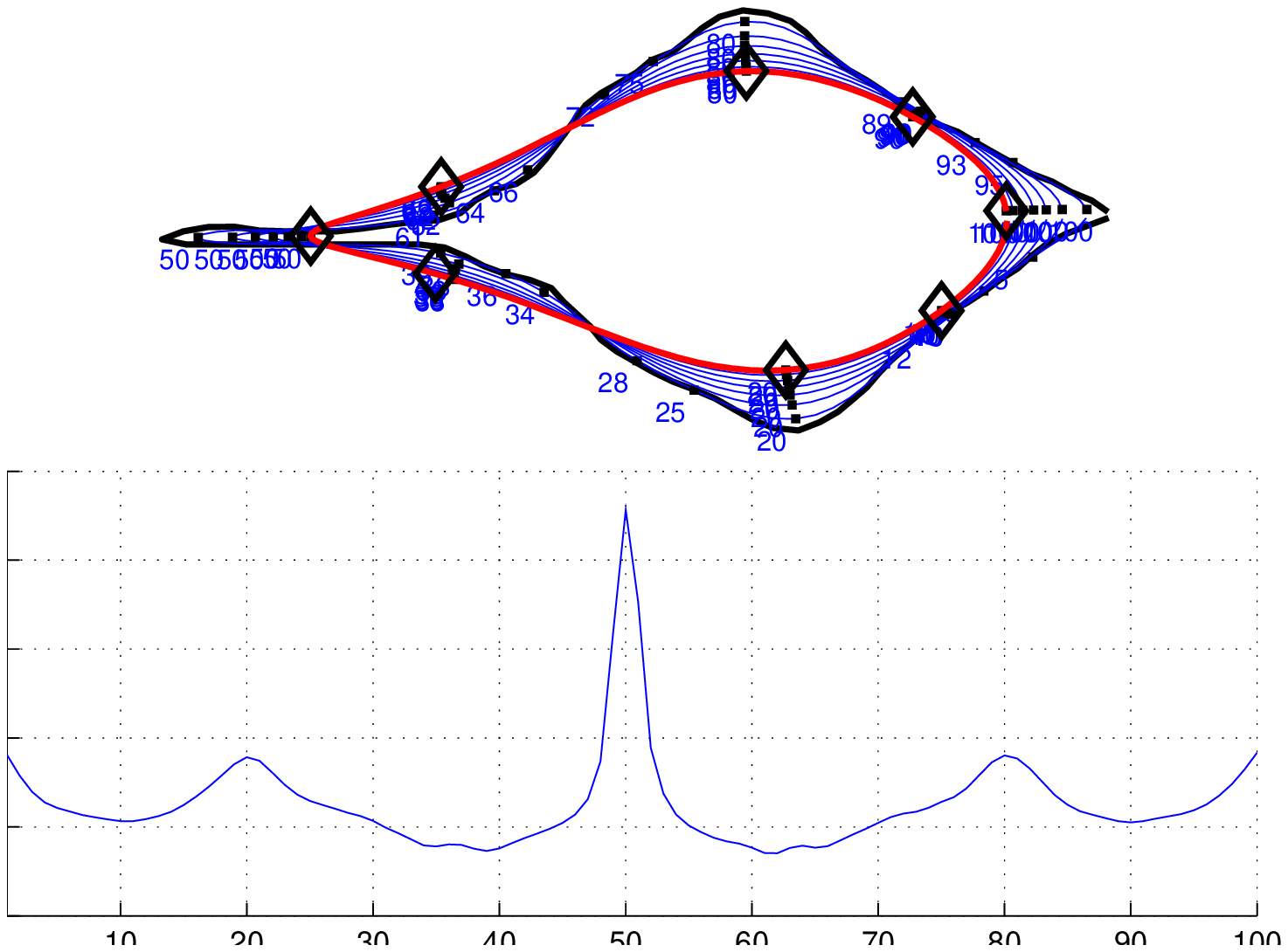}
\end{minipage}
\begin{minipage}[b]{\m cm}
\includegraphics[width=\w cm, height=\h cm]{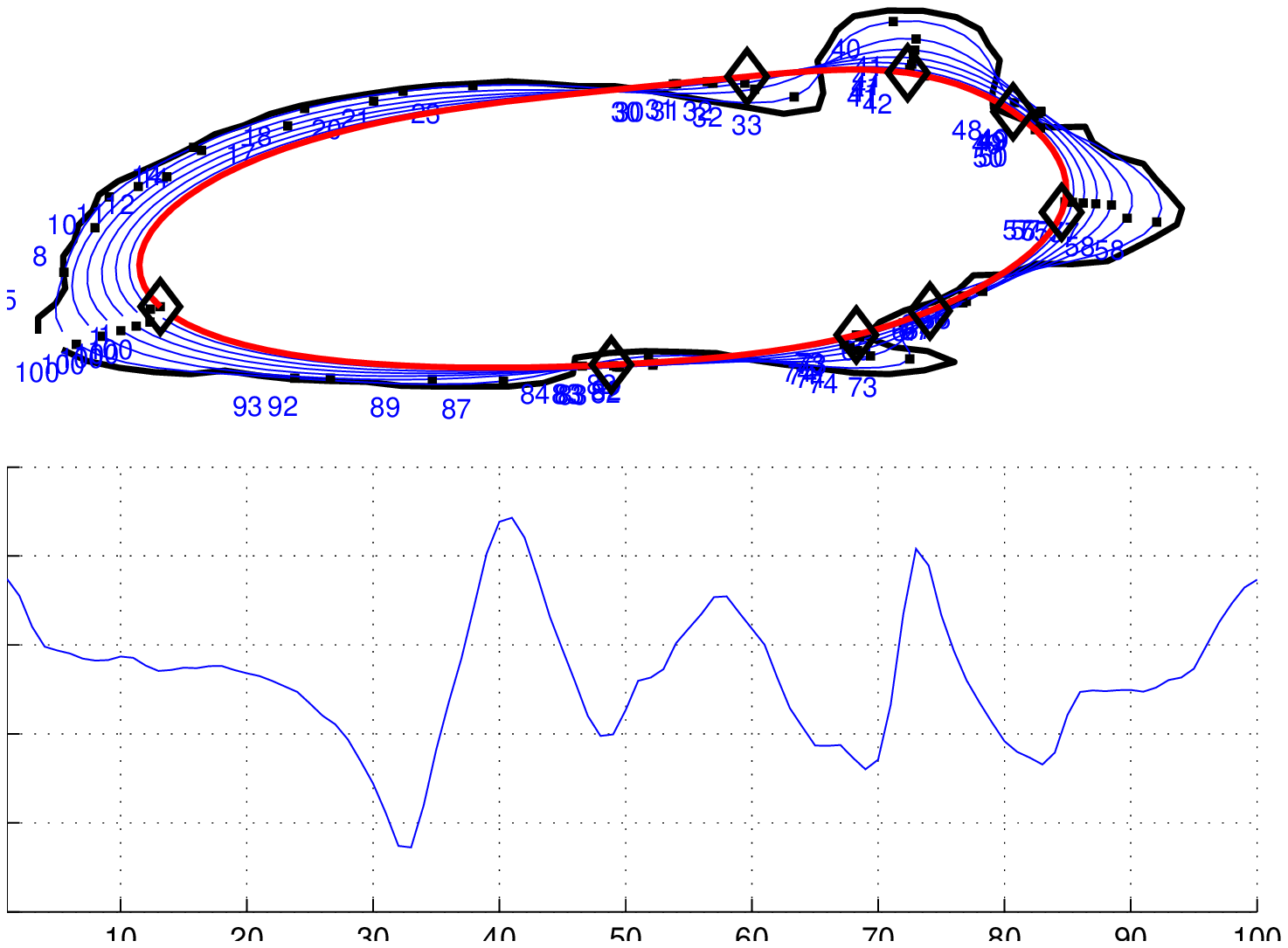}
\end{minipage}
\begin{minipage}[b]{\m cm}
\includegraphics[width=\w cm, height=\h cm]{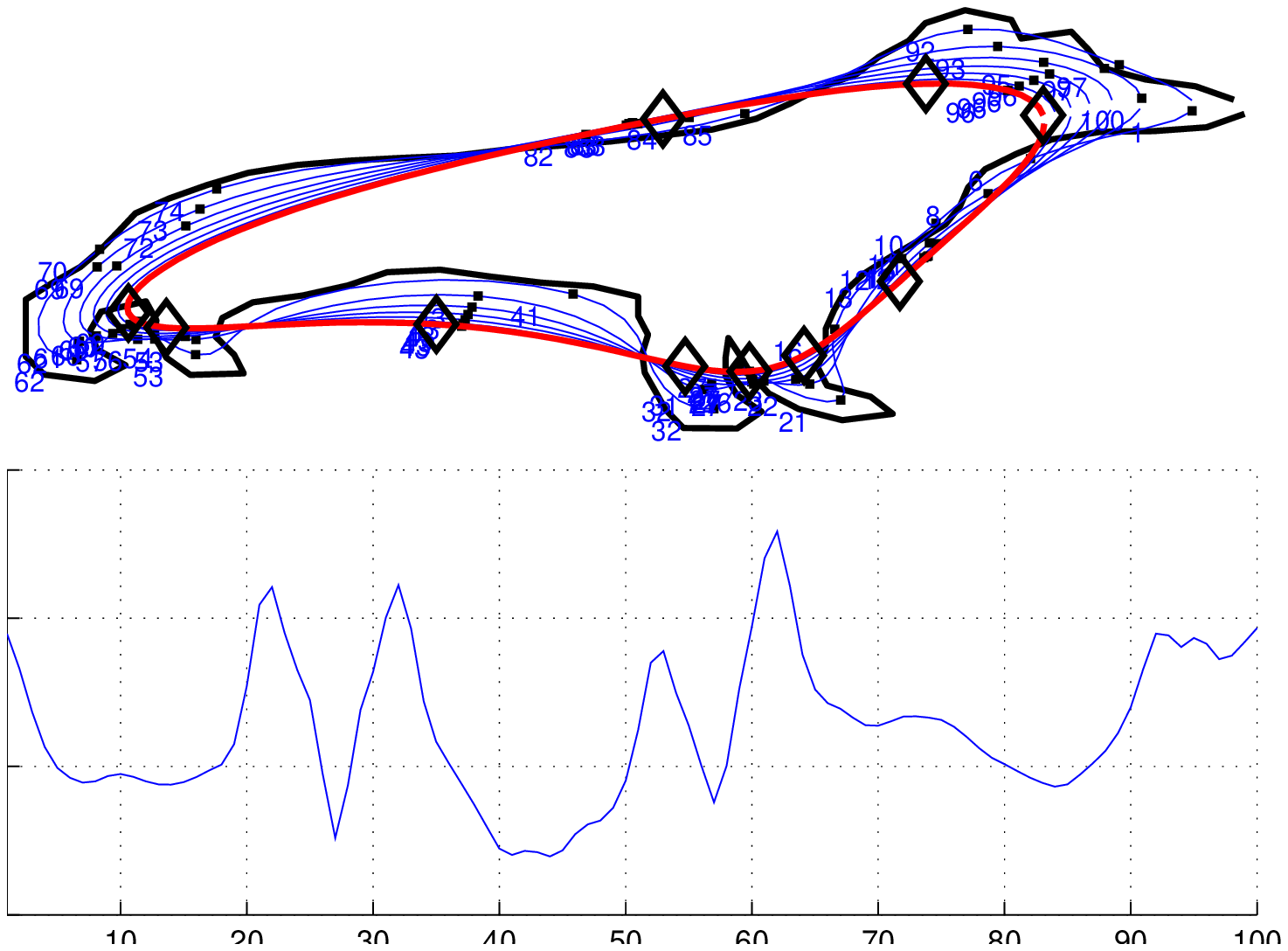}
\end{minipage}
}

\caption{Calculation of the ground truth (GT) set of Interesting Points (IP) of representative KIMIA silhouettes from all classes. The initial curve is shown with the thick black outermost contour, the final curve is shown as the red innermost contour. Intermediate curves produced by progressively smoothing the initial curve are shown as blue contours. Solid squares are tracking the local extrema of curvature at each smoothing step. The final ground truth points are shown as black diamonds. These are the local extrema of the cumulative curvature shown at the respective lower halfs.}
\label{fig:css}
\end{figure*}

\subsection{Ground Truth}\label{sec:GT}
We define the ground truth (GT) as the set of points where the \textit{cumulative curvature} function achieves local maximum or local minimum. The \textit{cumulative curvature} for a certain shape is defined as the point-wise addition of the curvatures of its progressively smoothed versions. In Fig.\ref{fig:css} the cumulative curvature function is shown together with the progressive smoothed versions of KIMIA silhouettes from all classes. The interesting points detected at each smoothing step are also shown as the local extrema of the respective curvature functions (vertices). The technique of tracking the persistence of special points across repeated smoothings, is a generally accepted method for identifying salient points, one of its variants known as Curvature Scale Space \cite{Abbasi99}, is an MPEG7 standard and in use for many years serving a wide range of applications in computer vision. GT Interesting Points (IPs) are calculated as in Fig.\ref{fig:css} only for the original KIMIA silhouettes of 100 points and they are assumed to their respective locations when the number of contour points are changed with \textit{noising}. If e.g. point No $n, 1\le n\le100$ was designated as a GT IP on the original 100 silhouette contour, then point No $2^p(n-1)+1$, where $p=log_2(N/100)$, is also a GT IP of the noisy contour of $N$ points, $N=100, 200, 400, 800, 1600$.   
This definition of GT IPs as consistent vertices across scales is generally accepted, as corners and high curvature points are traditionally considered \textit{interesting points}. Tracking vertices across scales combines both local (vertices) and global (scaling) characteristics of the shape in an perceptually objective and generally accepted manner. This algorithm of defining GT IPs is denoted by $SK$ (from Sum of Curvatures (K)) and is included in the experiments as the method one would expect to be the most successful in tracking back these GT points from the noisy versions. As we will discuss later however,  the proposed method achieves better Precision/Recall (PR) performance in tracking back these points, even though $SK$ is the method that defines them in the first place.  

\subsection{Methods Under Comparison}\label{sec:methods}
Colors and markers below refer to Fig.(\ref{fig:pr1b}). 
\begin{itemize}
\item
\textbf{Local Area Integral Invariant ($AI$)}: Using a circular kernel (constructed as a binary image of a circle of radius 15, as is suggested in \cite{ssoatto}) we convolve the filter with the shape image only at the boundary points. The values of the convolution at each of the boundary points are the values of the $AI$ estimated curvature at these points. As IP we pick the points where the $AI$ descriptor attains a local minimum or local maximum. Complexity: $O(k^2\times n)=O(n)$, $k$ being the size of the kernel and $n$ the number of contour points + at least one order if dynamic adjustment of the kernel size is needed in noisy conditions.  
Parameters: Circle Radius: 15, Color: red, Marker: star.
\item
\textbf{Proposed, 1st implementaion} ($V_o$):
In the first GL equation (\ref{eq:L156}) the first derivative of $\varphi$ is calculated as an integral over parametric angles, no distances are computed either and there is no derivative involved. For the third derivative of $\varphi$, a similar calculation is not apparent from the theory. For this reason we decided to provide two implementations of the proposed method. The first, signified by $V_o$ uses  the zero crossings of $\dot\varphi$ only and identifies IPs there without calculating derivatives. 
Complexity: $O(n^2)$ regardless of noise. Color: blue, Marker: Circle. 

\item
\textbf{Proposed,} $2^{\text{nd}}$ \textbf{implementation}($V$):
The second implementation signified by $V$ uses the collocation of the first and third derivatives. For the calculation of $\tdot\varphi$ see the relevant section \ref{sec:EXP}.
Complexity: $O(n^2)$ regardless of noise. Color: Green, Marker Cross.

\item
\textbf{Heron Curvature} ($K$): For each contour point $p_i$ a triangle is defined having vertices $p_i,p_{i-k},p_{i+k}$.  Heron Curvature at $p_i$ is defined as the area of this triangle. IPs are considered the local extrema of Heron Curvature. 
Complexity: $O(n)$ regardless of noise. Color cyan, Marker none.  

\item
\textbf{Cumulative Curvature} ($SK$): Same with cumulative curvature for defining GT IPs above (section \ref{sec:GT}) with the only difference that $SK$ will run on the noisy contours.
Complexity: $O(n^2)$ regardless of noise. The number of smoothings for calculating the cumulative curvature are a function of the number of points. Color magenta, Marker square.  
\end{itemize}

The methods under comparison were chosen to be pure approaches that emphasize different degrees of locality in solving the problem of IP detection, thus the investigation could remain focused on the essential concepts (e.g. what degree of locality is involved in the concept of an IP under the presence of noise?), rather than dragged into heuristic implementations directed to specific datasets or specialized problems. $K$ is a local method binded to a 1D locality, defined as a portion of the contour length, $AI$ binds to the locality of a 2D disk of a certain radius, whereas the proposed method is globally defined ($\dot\varphi$ defined through an integral) and finally $SK$ is a hybrid method since both the vertices (local) but also the tracking of them across scales (global) are used by the method.

\subsection{Performance Metrics}\label{sec:PR}
Every shape will in general have different number of ground truth (GT) points and each method under comparison will in general identify a different number of interesting points (IP) for such each shape. The challenge therefore in designing a strategy for comparison is in defining precision and recall in such a way that will not bias in favor of a particular method. We can hardly do better than using a standard probabilistic framework to calculate probability densities of a point being an IP under a particular method. 
For each method and for each point on the boundary we calculate the probability of this point to be an IP under this method as the reciprocal of its boundary distance to the nearest IP point predicted by this method. Here we imply a uniform distribution which is a plausible and unbiased assumption. Suppose therefore that $L_m=\{l_i\}, i=1,\dots ,n$ is the set of the $n$ IPs predicted by method $m$. For each point $p_j$ on the shape's boundary the probability of $p_j$ being an IP under method $m$ according to the uniform distribution is:  
\begin{equation}
\label{eq:exp1}
\eqalign{
pr(p_j \in IP)=ds  (\min_{i}(|p_j-l_i|))^{-1}
}
\end{equation}
\noindent
where $ds$ the arc element. This is indeed so since $(\min_{i}(|p_j-l_i|))^{-1}$ is the probability density for the uniform distribution in the interval $(p_j,l_i)$ if $l_i$ is the closest IP to $p_j$, identified by method $m$. The probability in other words, of a noisy point $p_j$ (point on a noisy version of the curve) being an IP at the original smooth version of the same curve according to a particular method, is inversely proportional to its distance to the nearest of the IPs identified by this method on the noise curve. As is typical with probability densities we assume $l_i \ne p_j$, their distance in other words is never zero but achieves a minimum value. 

Each method under comparison therefore, defines a probability distribution over points on a noisy curve that measures their potential to be the IPs of the original curve. But since the actual IPs are provided in the set of ground truth (GT) points as above, the methods will be compared on the cumulative probability they assign to the points that lie at the same location with the GT points. 
\begin{figure*}[]
\begin{center}

\includegraphics[height=13cm]{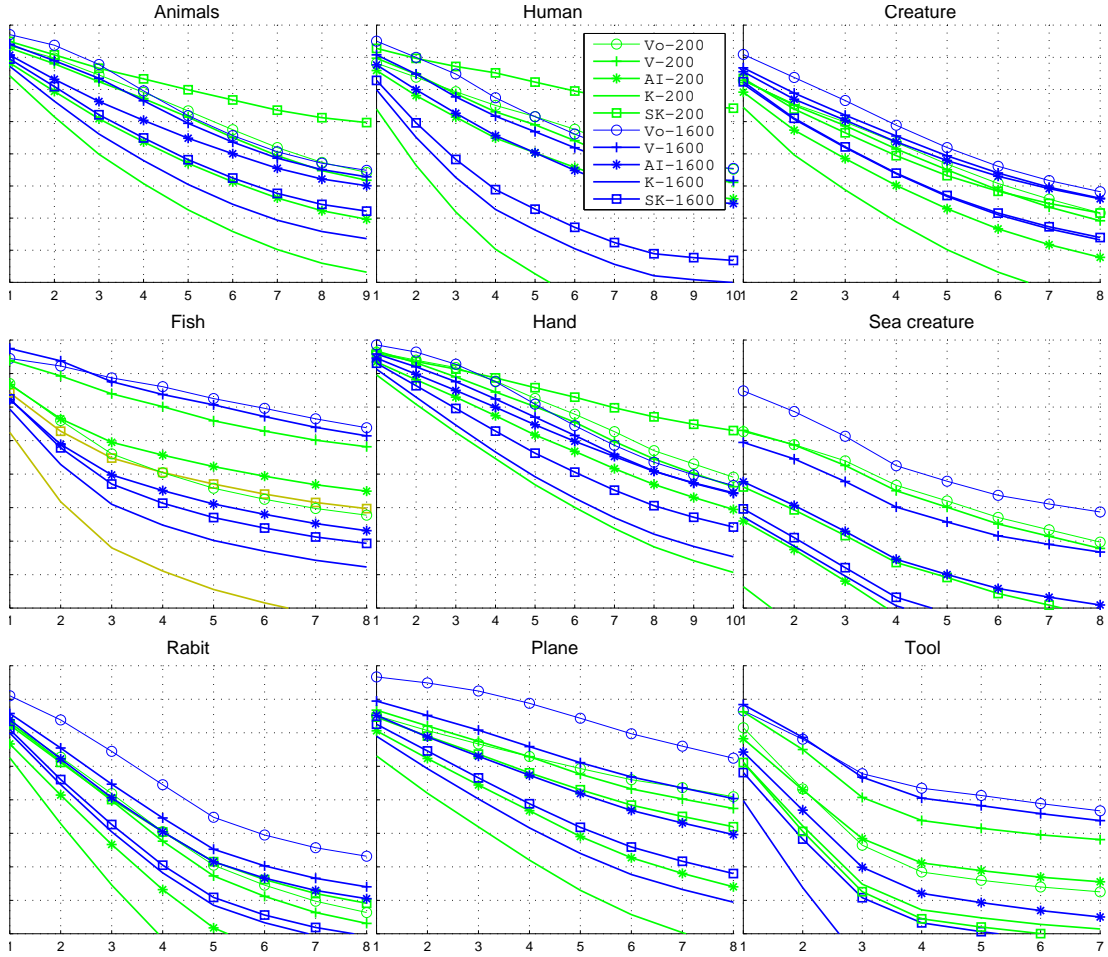}
\caption{Precision-Recall diagrams for each class of the KIMIA benchmark dataset of silhouettes. Blue graphs correspond to 1600 point contours, green graphs to 200 point contours. The PR graphs of all the 11 shapes of each class where averaged and presented here for each noising scenario. Horizontal axis is indexed with the IPs for each shape, vertical axis holds the precision in the range [0,1] but is omitted for better display. See section \ref{sec:results} for a discussion on the results.}
\label{fig:allcats}
\end{center}
\end{figure*}
 
A concern when designing the experiment was that local methods under noisy conditions typically identify IPs everywhere, densely at many locations, therefore IPs are randomly detected close to GT points. The probability framework just introduced compensates for this case since normalizing the probability density will distribute the probability mass evenly among all the points with low values at each point.  When on the other hand, a method identifies fewer IPs, normalizing the probability density will produce picks of high probability mass at these points and low probability mass at the other points. Identifying a correct location close to GTs is thus more critical in the latter cases. The probabilistic framework therefore is consistent with our intuition that a trade-off should exist between the number and the importance of location of the points identified by each method.

PR graphs are usually considered in a discrete context to measure the quality of matching in benchmarked  classification tasks \cite{Manning2008}. If a method, in order to get a correct match on the $m$th GT item, has previously matched correctly only $n, n\le m$ items, then the respective PR graph holds at the recall position $m$, ($m$ the items that have been recalled so far), the value $n\over m$, the ratio of the correct items over all items that were classified so far. In our experiment such a discrete approach was not apparent since there are different GT points to be recalled by methods that produce a different number of IPs. However, the probabilistic approach just introduced, can be seen as extending the discrete PR concept above, to continuous values, where instead of \textit{hit or no hit} we have the \textit{degree of a hit}, measured by the diversion of the method's density from the GT density at the location of the GT points. 

Since each method produces a probability density over all the points on the curve, this density can be compared to the GT density, produced by the GT method also over all curve points. This way, a common reference was made possible. The PR graph therefore, holds at the recall position $m$ (number of GT points recalled so far) the value $1-\sum_{i=1,\dots m}\vert p(i)-p_G(i)\vert$, where $p(i)$ and $p_G(i)$ are the method's density and the GT density respectively evaluated at the GT point $i$, and $\sum_{i=1,\dots m}\vert p(i)-p_G(i)\vert$ is the cumulative absolute diversion this method's density has from the GT density at the points recalled so far. We subtract from 1 to be visually consistent with the usual image of a PR graph that drops with precision.  For each density, the absolute differences of its values from the GT density values at the GT locations are sorted in ascending order and subtracted from 1 to be consistent with the Precision-Recall (PR) formulation. The final PR graph is a cumulative progressive addition of these sorted values. A perfect match will produce zero differences from the GT density at the GT locations (the method's density will be identical to the GT density in this case), therefore the visualized Precision Recall graph will be a constant 1 at all the GT locations. The best matches (smaller absolute differences from the GT density values) are sorted first. The PR graph drops as more GT IPs are examined and more errors are accumulated. 

\subsection{Implementation Details}\label{sec:EXP}
Local extrema and zero crossings are calculated for scalar descriptors using a level set approach.  A sliding window (1D window as a portion of the boundary in the form of $[s-ds,s+ds]$, where $s$ contour length parameter) is used on the values of the descriptor. The values left and right of the window center are subtracted from the center value in pairs and a local extreme is identified if all these differences have pairwise the same sign. The size of this window is the same for all methods and equals 0.017 as a ratio of the total contour length, therefore is invariant to the number of contour points. All local extrema calculations for all methods under comparison are calculated in this way using  the same window size ratio. The experiment is now described in steps:
    
\begin{enumerate}
\item
Each silhouette in the KIMIA dataset is discretized by 100 equally spaced points and the ground truth GT set of IPs is calculated.
\item A 100-point noisy version of each silhouette is constructed. Gaussian noise is applied by moving each point on the direction of its normal by a random quantity drawn by the normal distribution with variance 2.0. 

\item Noising is applied on each of the 99 100-points noisy silhouettes of the previous step producing 99 200-points noisy silhouettes. Similarly contours of 400, 800 and 1600 points are produced as explained in section \ref{sec:ED}. Each noising step adds a new point at the middle of each edge with a constant normal perturbation equal to $0.01$ times the magnitude of that edge.  

\item The 5 methods of \ref{sec:methods} are used to compute interesting points on each of the noisy silhouettes of the previous step.  In method $V$, for the existence of the $\tdot\varphi$ zero crossing a level set approach is applied: Since the locations of the local extrema of $\varphi$ have been identified on the zero crossings of $\dot\varphi$, one can infer the behavior of the third derivative of $\varphi$  by examining the \textit{shape} of $\varphi$ \textit{around} those points following a level set approach (since $\varphi$ is a scalar). A zero crossing of the third derivative in the vicinity of these zero crossings means a local extreme for the second derivative of $\varphi$ and thus a sudden change (above a threshold) in the values of $\varphi$ around its local extreme. The same sliding window strategy of the same size as above is used but now at least one of the pairwise absolute differences from the center must also be greater than 0.15 times the window length. $SK$ is the only method that smooths the curve and it does so progressively, it is thus the same method that calculates the GT set initially, but now works on the noisy curves after noising.  
  
\item For each noisy contour (after noising) and for each method, the probability densities are estimated according the section \ref{sec:PR} based on the IP set predicted for each contour by each method.

\item The comparison of the various densities (methods) is performed against the \textit{GT density} as explained in section \ref{sec:PR}. 
\end{enumerate}

The execution time for one contour of 100 points was less than a second on a computer with standard configuration for all methods. All methods had similar performance as contour points increased. The worst performance was observed in relation to the $SK$ method since multiple smoothings increased computational time in contours that exceeded 1600 points. 

\newcommand*{\hh}{15}
\newcommand*{\mm}{8.6}
\newcommand*{\mmm}{9.5}

\begin{figure*}[]
\makebox[\textwidth][c]{
\begin{minipage}[b]{\mm cm}
\includegraphics[height=\hh cm]{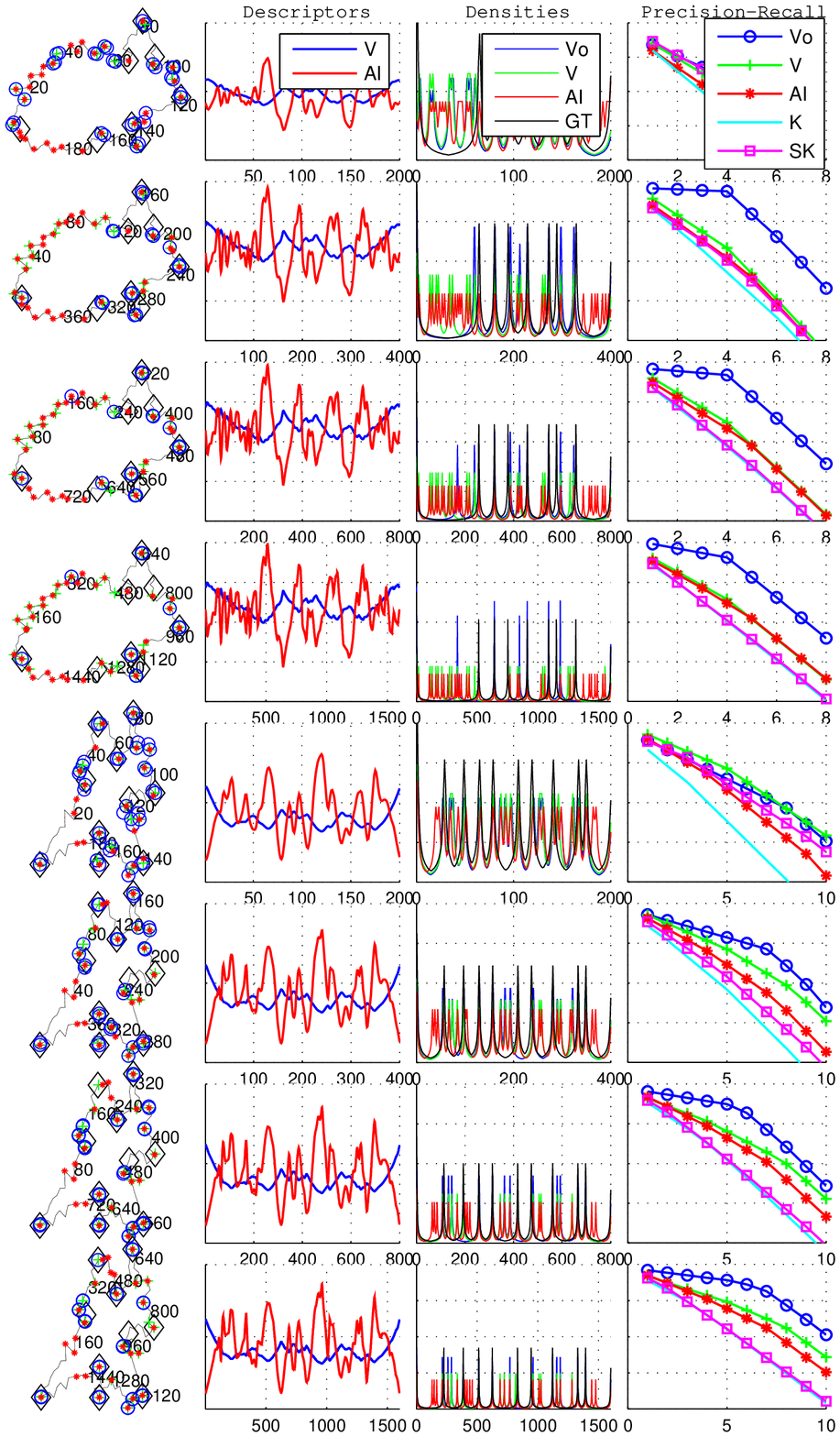}
\end{minipage}
\begin{minipage}[b]{\mmm cm}
\includegraphics[height=\hh cm]{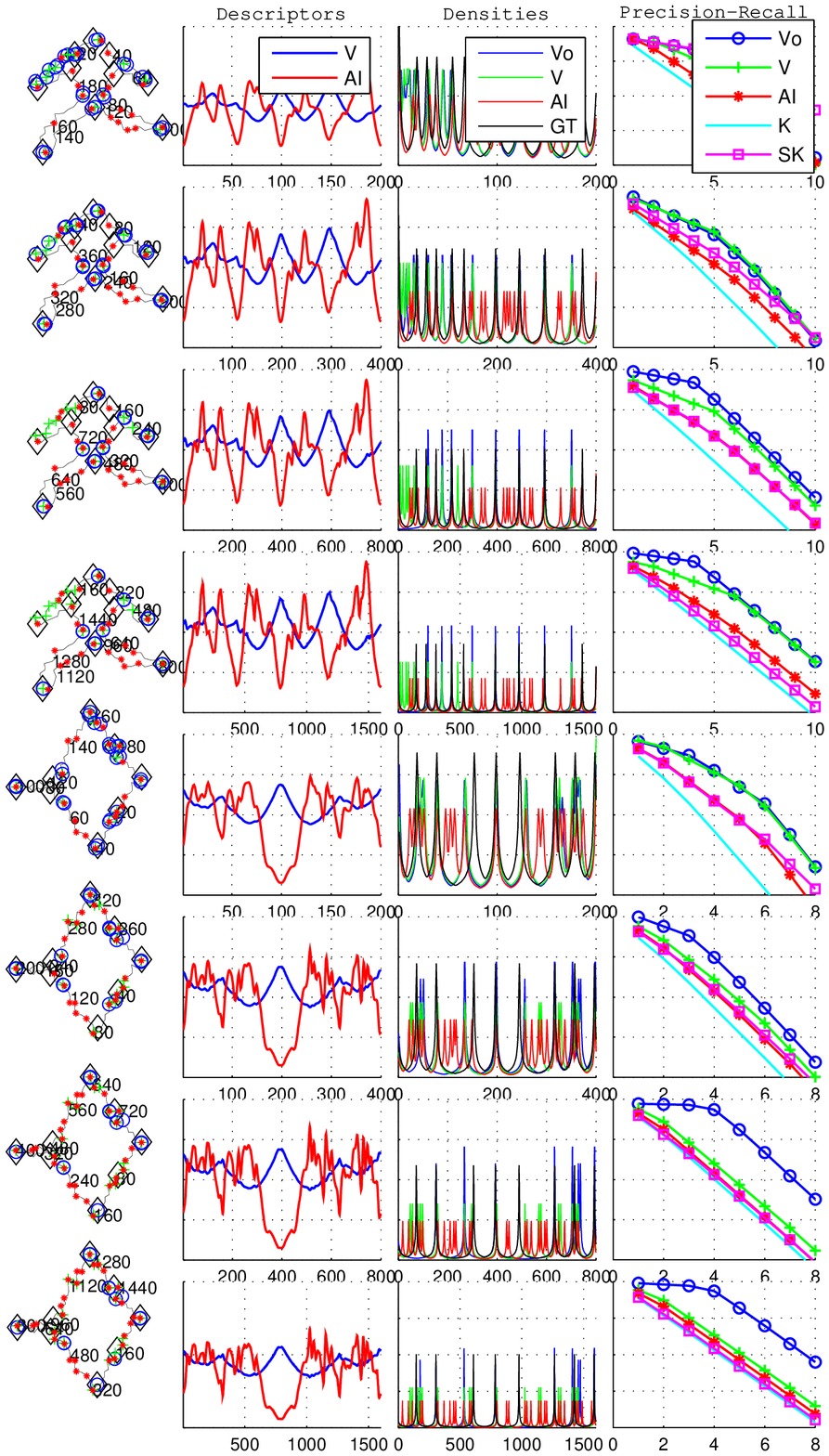}
\end{minipage}
}
\caption{$1^{\text{st}}$ column: Noisy versions of KIMIA shapes from representative classes in quartets of increasing \textit{noising} of 200, 400, 800 and 1600 points in vertical order. Big black diamonds represent the Ground Truth (GT) points identified by the cumulative curvature method as in Fig.(\ref{fig:css}) on the corresponding smooth silhouettes of 100 points. The positions of the GT points on the noisy shapes were extrapolated from those of the smooth shapes. GT points were calculated only for the initial KIMIA silhouettes of 100 points. Blue circles, green crosses and red stars represent Interesting Points (IPs) identified by $V_o$, $V$ and $AI$ methods respectively. $2^{\text{nd}}$ column: $AI$ and VAR ($\varphi$) descriptors, where the x-axis is marked by the number of contour points and the y-axis has been omitted for better display, since only the \textit{locations} of local extrema are important in this graph. $3^{\text{d}}$ column: Probability densities corresponding to $V_o$, $V$, $AI$ and Ground Truth $GT$ descriptors where the x-axis is marked by the number of contour points. The y-axis range is $[0,1]$ but was omitted for better display. $4^{\text{th}}$ column: Precision-Recall (PR) graphs for the 5 methods under comparison where the x-axis marks the GT points (the cardinality may differ among shapes) and the y-axis range is $[0,1]$ but was omitted for better display. The PR graph for a method, accumulates the diversion this method's density has from the GT probability density as more GT points are recalled. See Section \ref{sec:results} for a discussion on the results and Fig.\ref{fig:allcats} for per class average PR graphs.}
\label{fig:pr1b}
\end{figure*} 

\begin{figure*}[]
\makebox[\textwidth][c]{
\begin{minipage}[b]{\mm cm}
\includegraphics[height=\hh cm]{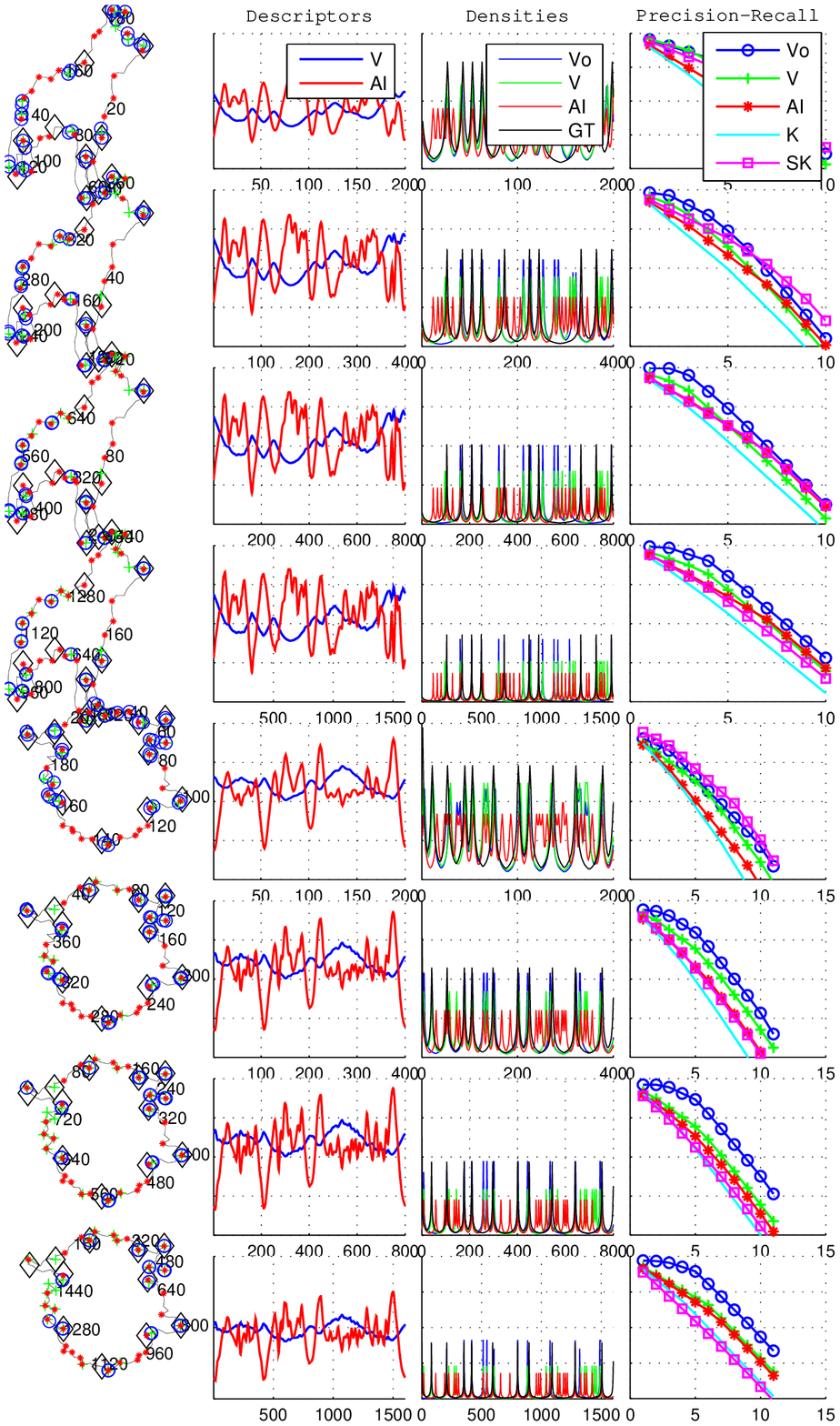}
\end{minipage}
\begin{minipage}[b]{\mmm cm}
\includegraphics[height=\hh cm]{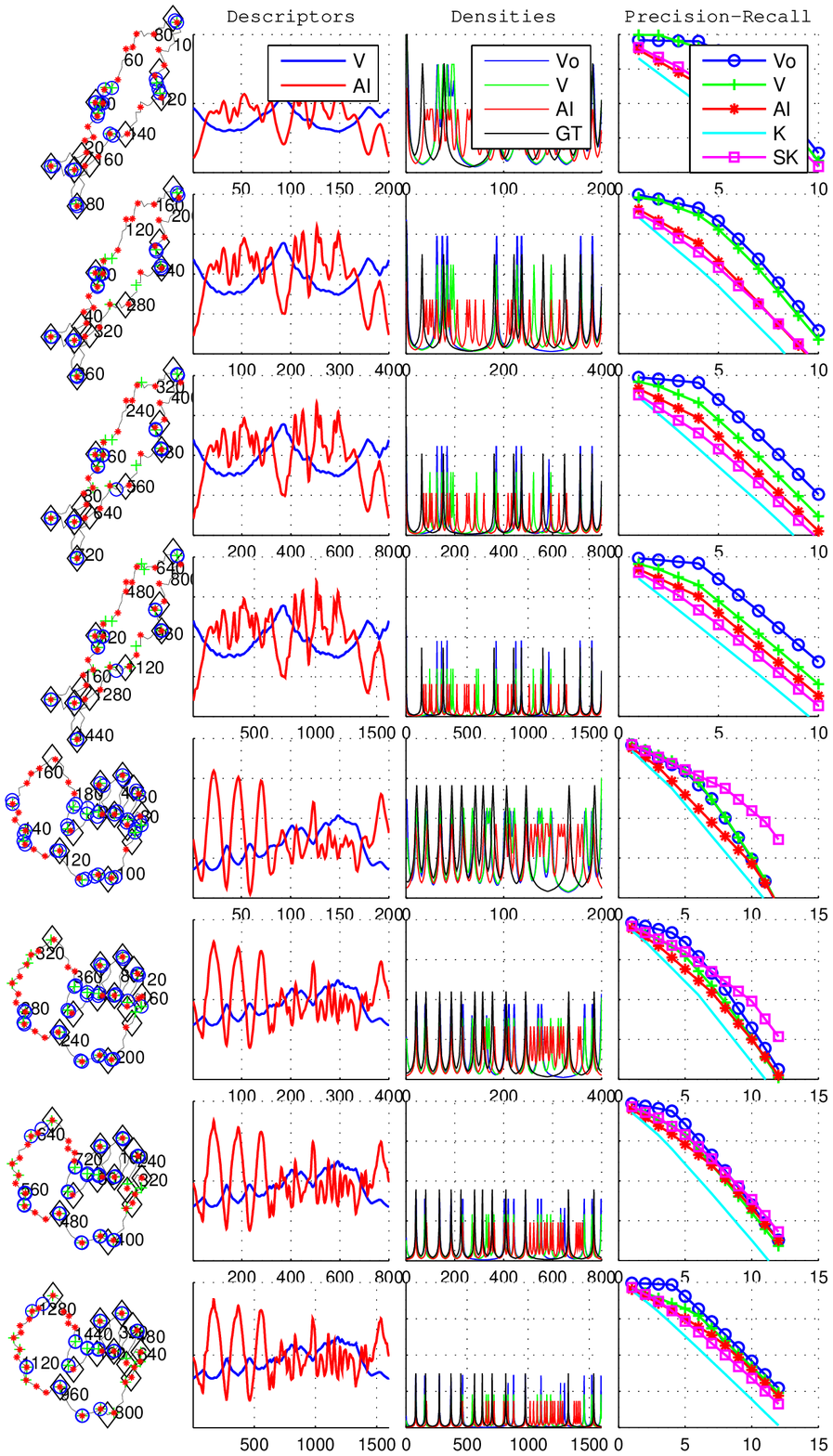}
\end{minipage}
}
\caption{Caption as in Fig.\ref{fig:pr1b} above.}
\label{fig:pr1c}
\end{figure*}

\section{Results}\label{sec:results}
In Fig.(\ref{fig:allcats}) PR measurements are presented per KIMIA shape class for all methods and for two noising scenarios. Green curves correspond to shapes affected by a 2 step noising (200-points noisy shapes) while blue curves correspond to a 5 step noising scenario (1600-points noisy shapes). See legend for the markers used for each particular method. In Fig. (\ref{fig:pr1b}) details of the process is shown for characteristic shapes of 2 KIMIA classes. See caption for details on the presentation.
Since performance is judged against GT IPs identified by the $SK$ method one would naturally expect $SK$ to perform extremely well in this set of experiments. We see in both of the above figures that $SK$ is affected by noising although it still performs well especially for shapes that have many dominant IPs of positive and negative curvatures. Even in those cases however, noising improves the proposed method $V_o$, especially for the most interesting points (best matches).   

The proposed $V_o$ is better than all the other methods, the superiority gap, seen as the vertical distance of PR curves, increases with noising. Compared to $SK$, $V_o$, behaves better in most shape classes. For certain classes of shapes with many dominant points of positive and negative curvatures (Hands, Humans, Animals) $SK$ identifies GT IPs more accurately for low noising scenarios but as noising increases the performance of $SK$ deteriorates, while $V_o$ improves to a point where it outperforms $SK$ for the best 2-3 interesting points. It is important to notice for one more time that $SK$ is a biased estimator in this experiment. 

We also notice that both implementations of the proposed method improve with noising. As noising increases the proposed method identifies less points in contrast to the other methods where more points are identified as noising is recursively applied. This improvement at the PR graphs is particularly apparent for the first 3,4 best interesting points, and this is easily explainable since less but more accurate points are correctly identified after noising.  
This behavior can be observed e.g. in Fig.(\ref{fig:pr1b}) airplane class, where the PR graph improves with noising to an almost perfect score for the best 5 hits at the 4th noising step (1600points), while the score at the 2d noising step (200 points) was comparable to the other methods. Similar behavior, where the best 4-5 hits approach a perfect score with noising can be observed for the rest of the classes of the same figure, and eventually for all the KIMIA classes from Fig. (\ref{fig:allcats}) where the scores from all shapes per class are combined. What can be observed for the $AI$ and $K$ methods is that they are not affected significantly with noising, something that was expected since noising does not change the overall shape significantly. This reminds us that the whole purpose of noising is to enrich the tangent directions around the curve points and this can be done effectively by introducing slight perturbations with a minimum effect to the appearance of the shape.   

\begin{figure}[t]
\begin{center}
\includegraphics[height=8cm]{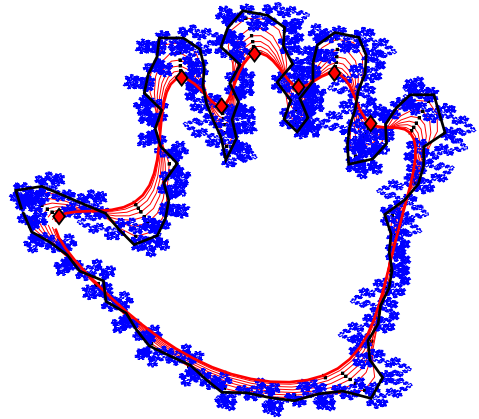}
\end{center}
\caption{Space filling properties of incremental noising and progressive smoothing. The black curve is the initial contour, red curves are the progressive smoothing versions of the initial contour, thick red line is the final smoothed version of the initial contour, red diamonds are the persistent vertices identified by both methods. The blue curve is the initial curve after 10 levels of noising ($2^{10}100$-points contour) filling the space around the initial contour in a fractal manner. Both progressive smoothing and incremental noising detect persistent vertices at places of increased 2D-space coverage.}
\label{fig:fractal}
\end{figure}  
\section{Discussion}

In the experiment above, a surprising result is obtained. On distorted shapes, incremental noising performs better in identifying Ground Truth (GT) points that were defined by progressive smoothing on the corresponding undistorted shapes. This can happen only if: (a) The two methods are effectively identifying the same vertices, and (b) incremental noising is more robust to Gaussian shape distortions than progressive smoothing. Even though the two methods are conceptually orthogonal there must be a link that connects them in the context of vertex identification. In this direction, one may notice that the shape features that are persistent in the course of progressive smoothing are also the ones that are emphasized by the \textit{space filling} properties of incremental noising.

Indeed, the successive contours produced by progressive smoothing \textit{fill up} the 2D space between the initial contour and the final smooth version. Persistent vertices are identified in places where the intermediate contours shrink faster per single smoothing, thus the local space coverage is increased. Incremental noising on the other hand, adds more points to the initial contour at each step, producing a curve that covers 2D space around the original contour in a \textit{fractal} manner (blue colored space filling curve in Fig.(\ref{fig:fractal})). As one can see in the same figure, persistent vertices are located in areas where the 2D-space filling curve is more dense and these are the same areas where progressively smoothed contours cover more ground as well. The \textit{fractal} curve produced due to incremental noising will \textit{fill up} more area around strong vertices than it will around weaker vertices and this is why stronger vertices persist as well with noising as they do with smoothing. Incremental noising works because it is used in combination with VAR's global characteristics. As has been already explained, VAR is sensitive to relative location on the curve and less to noise-like curvature. The fractal curve produced by incremental noising evolves in a way that eliminates location specific characteristics faster around weak vertices than around stronger ones.  
In smooth shapes therefore, the two methods identify the same points, after distorting the shapes however, incremental noising seems more robust in tracking back those same points, exhibiting a noise removal behavior in that sense.
But this is not the only advantage of incremental noising in this task. Smoothing is a lossy operation whereas noising is not as only new points are always added to the previous ones. A sub-sampling can always produce the initial curve. Furthermore, the space filling coverage due to incremental noising converges to a certain area \textit{around} the original contour that depends on the initial noising parameter. Progressive smoothing on the other hand always covers the same area, namely the one between the initial contour and the final smooth version, the space covering behavior is thus more discriminative and better controlled in the case of incremental noising.    

\bibliographystyle{splncs}
\bibliography{egbib}

\begin{thebibliography}{10}

\bibitem{CVPR14}
Raftopoulos, K., Ferecatu, M.:
\newblock Noising versus smoothing for vertex identification in unknown shapes.
\newblock In: Computer Vision and Pattern Recognition (CVPR), 2014 IEEE
  Conference on. (June 2014)  4162--4168

\bibitem{CVIU}
Raftopoulos, K., Kollias, S.:
\newblock The global~--~local transformation for noise resistant shape
  representation.
\newblock Computer Vision and Image Understanding \textbf{115}(8) (2011)  1170
  --1186

\bibitem{ssoatto}
Manay, S., Cremers, D., Hong, B.W., Yezzi, A., Soatto, S.:
\newblock Integral invariant signatures for shape matching.
\newblock Pattern Analysis and Machine Intelligence, IEEE Transactions on
  \textbf{27}(11) (2006)  1602--1618

\bibitem{rel1}
Bennett, J.R., Mac~Donald, J.S.:
\newblock On the measurement of curvature in a quantized environment.
\newblock IEEE Trans. Comput. \textbf{24}(8) (August 1975)  803--820

\bibitem{rel2}
Pottmann, H., Wallner, J., Huang, Q.X., Yang, Y.L.:
\newblock Integral invariants for robust geometry processing.
\newblock Computer Aided Geometric Design \textbf{26}(1) (2009)  37 -- 60

\bibitem{rel3}
Michor, P.W., Mumford, D.:
\newblock Riemannian geometries on spaces of plane curves.
\newblock J. Eur. Math. Soc. (JEMS  1--48

\bibitem{rel4}
He, X., Yung, N.H.C.:
\newblock Curvature scale space corner detector with adaptive threshold and
  dynamic region of support.
\newblock In: Pattern Recognition, 2004. ICPR 2004. Proceedings of the 17th
  International Conference on. Volume~2. (2004)  791--794 Vol.2

\bibitem{rel5}
Magid, E., Soldea, O., Rivlin, E.:
\newblock A comparison of gaussian and mean curvature estimation methods on
  triangular meshes of range image data.
\newblock Computer Vision and Image Understanding \textbf{107}(3) (2007)  139
  -- 159

\bibitem{rel6}
Nguyen, T., Debled-Rennesson, I.:
\newblock Curvature estimation in noisy curves.
\newblock In Kropatsch, W., Kampel, M., Hanbury, A., eds.: Computer Analysis of
  Images and Patterns. Volume 4673 of Lecture Notes in Computer Science.
\newblock Springer Berlin Heidelberg (2007)  474--481

\bibitem{rel7}
Salmon, J.P., Debled-Rennesson, I., Wendling, L.:
\newblock A new method to detect arcs and segments from curvature profiles.
\newblock In: Pattern Recognition, 2006. ICPR 2006. 18th International
  Conference on. Volume~3. (2006)  387--390

\bibitem{rel8}
Delorme, M., Mazoyer, J., Tougne, L.:
\newblock Discrete parabolas and circles on 2d cellular automata.
\newblock Theor. Comput. Sci. \textbf{218}(2) (May 1999)  347--417

\bibitem{rel9}
Fleishman, S., Cohen-Or, D., Silva, C.T.:
\newblock Robust moving least-squares fitting with sharp features.
\newblock ACM Trans. Graph. \textbf{24}(3) (July 2005)  544--552

\bibitem{rel10}
Tong, W.S., Tang, C.K.:
\newblock Robust estimation of adaptive tensors of curvature by tensor voting.
\newblock Pattern Analysis and Machine Intelligence, IEEE Transactions on
  \textbf{27}(3) (2005)  434--449

\bibitem{rel11}
Daniels~II, J., Ochotta, T., Ha, L.K., Silva, C.T.:
\newblock Spline-based feature curves from point-sampled geometry.
\newblock The Visual Computer \textbf{24}(6) (2008)  449--462

\bibitem{rel12}
Calabi, E., Olver, P., Shakiban, C., Tannenbaum, A., Haker, S.:
\newblock Differential and numerically invariant signature curves applied to
  object recognition.
\newblock International Journal of Computer Vision \textbf{26}(2) (1998)
  107--135

\bibitem{rel13}
Tward, D.J., Ma, J., Miller, M.I., Younes, L.:
\newblock Robust diffeomorphic mapping via geodesically controlled active
  shapes.
\newblock Int. J. Biomedical Imaging \textbf{2013} (2013)

\bibitem{rel15}
Amit, Y., Geman, D., Fan, X.:
\newblock A coarse-to-fine strategy for multiclass shape detection.
\newblock IEEE Trans. Pattern Anal. Mach. Intell. \textbf{26}(12) (2004)
  1606--1621

\bibitem{kimia}
Sebastian, T.B., Klein, P.N., Kimia, B.B.:
\newblock Recognition of shapes by editing their shock graphs.
\newblock IEEE Trans. Pattern Anal. Mach. Intell. \textbf{26}(5) (2004)
  550--571

\bibitem{Abbasi99}
Abbasi, S., Mokhtarian, F., Kittler, J.:
\newblock Curvature scale space image in shape similarity retrieval.
\newblock Multimedia Syst. \textbf{7}(6) (November 1999)  467--476

\bibitem{Manning2008}
Manning, C.D., Raghavan, P., Schutze, H.:
\newblock Introduction to Information Retrieval.
\newblock Cambridge University Press, New York, NY, USA (2008)

\end{thebibliography}

\end{document}